\RequirePackage{fix-cm}
\documentclass[smallcondensed]{svjour3}     % onecolumn (ditto)
\smartqed  % flush right qed marks, e.g. at end of proof
\usepackage{graphicx} %use graph format
\usepackage{epstopdf}

\usepackage{amsmath}
\usepackage{amssymb}

\usepackage{subfig}
\usepackage{color,xcolor} %for colour

\usepackage[linesnumbered, ruled]{algorithm2e}
\SetKwRepeat{Do}{do}{while}%

\usepackage{verbatim}
\usepackage{float}
\usepackage{mathptmx}      % use Times fonts if available on your TeX system

\begin{document}

\title{Cascade Attentive Dropout for Weakly	Supervised Object Detection%\thanks{Grants or other notes
}


%\author{First Author         \and
%        Second Author %etc.
%}
%
%%\authorrunning{Short form of author list} % if too long for running head
%
%\institute{F. Author \at
%              first address \\
%              Tel.: +123-45-678910\\
%              Fax: +123-45-678910\\
%              \email{fauthor@example.com}           %  \\
%%             \emph{Present address:} of F. Author  %  if needed

\author{Wenlong Gao \and Ying Chen \and Yong Peng }


\institute{Wenlong Gao \at
              Key Laboratory of Advanced Process Control for Light Industry (Ministry of Education), Jiangnan University, Wuxi 214100, China  \\
              \email{gaowl@stu.jiangnan.edu.cn}           %  \\
           \and
           Ying Chen \at
			Key Laboratory of Advanced Process Control for Light Industry (Ministry of Education), Jiangnan University, Wuxi 214100, China  \\
           \email{chenying@jiangnan.edu.cn}
           \and
           Yong Peng \at
           Key Laboratory of Advanced Process Control for Light Industry (Ministry of Education), Jiangnan University, Wuxi 214100, China  \\
           \email{ypeng@jiangnan.edu.cn}
}

\date{Received: date / Accepted: date}

\maketitle
%--------------------
\begin{abstract}
Weakly supervised object detection (WSOD) aims to classify and locate objects with only image-level supervision. Many WSOD approaches adopt multiple instance learning as the initial model,  which is prone to converge to the most discriminative object regions  while ignoring the whole object, and therefore reduce the model detection performance. In this paper, a novel cascade attentive dropout strategy is proposed to alleviate the part domination problem, together with an improved global context module. We purposely discard attentive elements in both channel and space dimensions, and capture the inter-pixel and inter-channel dependencies to induce the model to better understand the global context. Extensive experiments have been conducted  on the challenging PASCAL VOC 2007 benchmarks, which achieve 49.8\% mAP and 66.0\% CorLoc, outperforming state-of-the-arts.
\keywords{Weakly supervised object detection \and Convolutional neural network \and Dropout}
\end{abstract}

%-------------------
\section{Introduction}
\label{intro}

\begin{figure*}[t]
	\centering{\includegraphics[width=0.9\columnwidth]{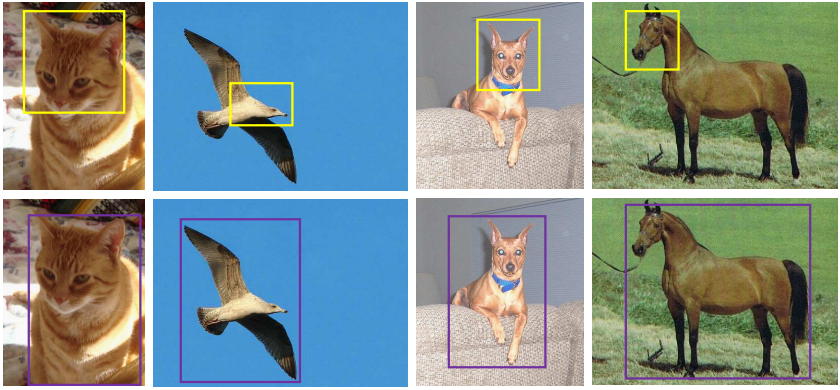}}%width=0.85\paperwidth
	\caption{Typical weakly-supervised object detection results: part-dominated (first cow)and correct localization (second row)}
	\label{fig:problem}
\end{figure*}

Weakly Supervised Object Detection (WSOD) is attracting more and more attention in computer vision area in recent years. The training of a traditional object detector usually is based on a large amount of manually labeled data, while the labeling process is time-consuming, expensive, and inefficient. Weakly supervised object detection has come into being, which aims to solve this problem by training a detector with only image-level annotations instead of bounding boxes.

Most methods model WSOD as a multiple instance learning (MIL) problem, where each image is considered as a bag and the object proposals as instances. During the training process, the network iterates in the following two steps: (1) training an object classifier to compute the object score of each object proposal; (2) selecting the proposals with the highest score and their similar proposals based on IoU. The model is prone to learn only the most discriminative object features rather than comprehensive object features so that it will cause part domination shown in Figure \ref{fig:problem}. For example, in the case of a cat, the head may be more discriminative than the body in which appearance changes dramatically due to patterns. In this case, previous techniques can localize only the head, rather than the entire region.

Recent work have alleviated this problem by using context information\cite{kantorov2016contextlocnet,wei2018ts2c}, progressive refinement\cite{tang2018pcl,tang2017multiple,wan2019c}, and smoothed loss functions\cite{wan2019c}. For example, Contextlocnet\cite{kantorov2016contextlocnet} built additive and contrastive guidance models to utilize their surrounding context feature to enhance localization performance. C-MIL\cite{wan2019c} divided the instances in the bag into multiple subsets, and defines corresponding smoother loss functions on the each subset to approximate the convex loss function.

Attention mechanism, which enables network to focus on the most informative views,  has been proven effective in many fields, such as image classification\cite{hu2018squeeze}, image inpainting\cite{yu2018generative}, medical image segmentation\cite{sinha2019multi}, etc. However, the mechanism hurts the object localization performance of WSOD because it focuses only on the most discriminative features. To address this issue, we propose a cascaded attentive dropout module (CADM) to inducing the network to learn less discriminative features for classification but meaningful features for object localization. Specifically, following channel attentive dropout, two complementary attentive branch are built and randomly selected to build spatial-attentive feature maps, where the one rewards the most discriminative features while  the other punishes them via a designed attentive dropout strategy. A global context module (GCM), which uses sigmoid to enhance nonlinearity and perform feature fusion through element-wise multiplication and additions, also been introduced into the proposed network to obtain better global context information.

In summary, the main contributions of our work can be summarized as follows:
\begin{enumerate}
\item[1)] An end-to-end weakly supervised object detection network is proposed, considering both network attention and global context information.
\item[2)] A lightweight but effective cascade attentive dropout module is designed to help the network learn more comprehensive features rather than only discriminative features, which notably improve the accuracy of WSOD.
\item[3)] An improved global context module is introduced to further boost the learned features in a more efficient way of feature fusion, jointly optimizing the region classification and localization.
\item[4)] The proposed network significantly outperforms most state-of-the-art weakly supervised object detection approaches on PASCAL VOC 2007.
\end{enumerate}

%-------------------
\section{Related Work}\label{sec2}

\subsection{Weakly supervised object detection}
Recent work have combined MIL and CNN to train a weakly supervised object detector in an end-to-end way. Bilen and Vedaldi\cite{bilen2016weakly} proposed a two-stream weakly supervised deep detection network (WSDDN) to get classification and detection scores respectively. Based on WSDDN, Tang et al.\cite{tang2017multiple} proposed an online instance classifier refinement (OICR) approach to refine the rough output of WSDDN, Kantorov et al.\cite{kantorov2016contextlocnet} introduced two different kinds of context-aware guidance to improve localization performance of the network. Tang et al.\cite{tang2018pcl} also proposed a graph-based center cluster method to alleviate the local optimum problem. Some work have also begun to link weak supervision and strong supervision to comprehensively solve the WSOD problem. Zhang et al.\cite{zhang2018w2f} designed a Weakly-supervised to fully-supervised framework (W2F) which mines high-quality pseudo ground truth to train a  fully-supervised object detector. Wang et al.\cite{wang2018collaborative} proposed a weakly supervised collaborative learning approach that adopts WSDDN and Faster-RCNN as weakly and strongly supervised sub-network respectively. From the perspective of optimization, Wang et al.\cite{wan2019c} introduce a continuation optimization method into MIL to boost the detection performance. Arun et al.\cite{arun2019dissimilarity} designed a novel dissimilarity coefficient based WSOD framework which is optimized by minimizing the difference between an annotation agnostic prediction distribution and an annotation aware conditional distribution. Besides, some work combined the tasks of weakly supervised object detection and segmentation into a unified end-to-end learning architecture\cite{gao2019c,zeng2019wsod2}.

\subsection{Dropout}
Dropout is a regularization technique originally proposed by \cite{hinton2012improving} to alleviate the over-fitting problem of fully connected neural networks, which randomly drops neurons with fixed probability during network training. However, all neuron activations are retained while testing the model, and the final output will be scaled according to the dropout probability. In this way, the network can easily integrate several different small models to achieve good regularization. Inspired by Dropout, Dropconnect\cite{wan2013regularization}, Monte Carlo dropout\cite{gal2016dropout}, and many others were also introduced. This is only useful for the fully connected layer, but useless for the convolution layer. One of the reasons may be that the spatially adjacent pixels on the feature map are related to each other and share contextual information. After that, DeVries et al.\cite{devries2017improved} designed Cutout to randomly drop out contiguous regions of input images, Tompson et al.\cite{tompson2015efficient} also proposed SpatialDropout to randomly drops partial channels of a feature map, rather than dropping the independent random pixels. Besides, ADL\cite{choe2019attention} was proposed to utilize the attention mechanism to erase the maximally activated part.
\vspace{-2mm}

\begin{figure*}[t]
	\centering{\includegraphics[scale=0.9]{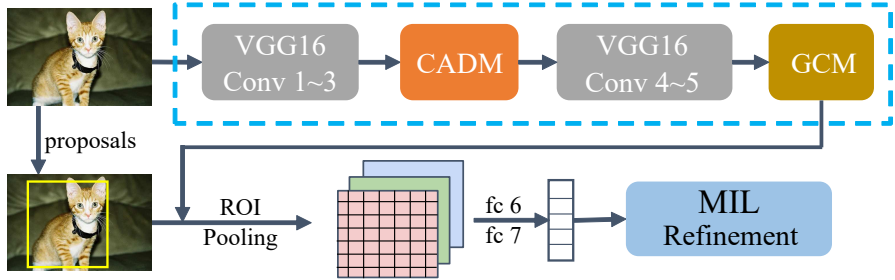}}
	\caption{Architecture of our proposed network. (1)Generate comprehensive feature maps by VGG16 with CADM and GCM. (2)Generate fixed-size ROI features. (3)Feed the proposal feature vectors to MIL and Refinement submodule to predict categories and locations.\label{fig:architeture_diagram}}
\end{figure*}

\subsection{Attention mechanism}
The Attention mechanism is inspired by the human vision which does not treat all data equally but enhances or weakens them. Recent work have been proposed to improve the localization performance of the model. For example, Hu et. al proposed a squeeze-and-excitation network (SENet)\cite{hu2018squeeze} to model the inter-channel dependencies, which generates a weight of $1 \times 1 \times C$ via a global average pooling layer and two FC layers and multiply it into the input feature map to get an enhanced feature map. Based on SENet, SKNet\cite{li2019selective} built multiple branches of different receptive fields and used the information summarized by the multiple scale feature to channel-wise guide how to allocate the representation of which kernel to focus on. Wang et. al proposed Non-local Neural networks to fuse the global information and bring richer semantic information to the following layers. Convolutional Block Attention Module (CBAM)\cite{woo2018cbam} are also proposed to  enhance features in channel and spatial dimensions in a cascading manner.

%-------------------
\section{The Proposed Approach}\label{sec3}
In this section, we will introduce our proposed weakly supervised object detection architecture. As shown in Figure \ref{fig:architeture_diagram}, the first stage aims to extract enhanced feature maps $\boldsymbol{X}_5^*$ from VGG16 with a cascade attentive dropout module (CADM) and a global context module (GCM). The enhanced feature maps and region proposals generated by Selective Search\cite{uijlings2013selective} are then sent to the RoI pooling layer to produce fixed-size RoI feature maps. At last, MIL Refinement Module utilizes proposal feature vectors $\boldsymbol{x}$ to predict object categories and locations. The proposed CADM, which is designed to elliminate negative effects of discriminative features, is employed on pooling 3 feature map. Different from ADL[21] which erased the maximally activated spatial parts, we purposely discard attentive elements in both channel and space dimension. The remainder of this section will discuss these components in detail.

\begin{figure*}[t]
	\centering{\includegraphics[scale=0.45]{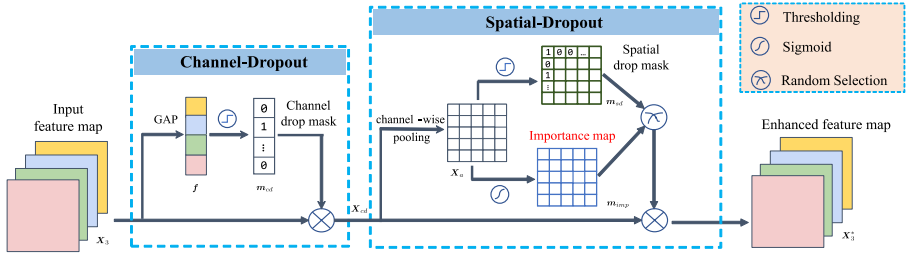}}
	\caption{Cascade attentive dropout module. This module dropouts elements in the dimensions of channel and space respectively to induce the model to learn more comprehensive features.
		\label{fig:cadm}}
\end{figure*}

\subsection{Cascade Attentive  Dropout Module}
Weakly supervised object detectors tend to learn only the most discriminative features in images\cite{wan2018min,wan2019c}, which will cause network localization errors and decrease detection accuracy. In order to solve this problem, we try to dropout elements in the dimensions of channel and space respectively, as shown in Figure \ref{fig:cadm}.

\textbf{Channel-Dropout}. Given a feature map $\boldsymbol{X}_3 \in \boldsymbol{R}^{N\times D\times H\times W}$ extracted from CNN, the channel-dropout module takes it as input and outputs a global information embedding via a global average pooling (GAP) layer. The embedding can also be considered as the confidence of different channels, denoted as $\boldsymbol{f}=(f_1,f_2,\cdots,f_D)$. Note that $N$ is the mini-batch number, $D$ is the number of channel, $W$ and $H$ are weight and height of the feature map, respectively. After that, we refer the confidence of the channel with the highest confidence as $f_{max}$, and set the threshold $\lambda_1$. When the channel confidence $f_i$ is greater than $f_{max} \cdot \lambda_1$, the channel $i$ is dropped; otherwise, keep the channel. Consequently, a binary channel-dropout mask $\boldsymbol{m}_{cd} \in \boldsymbol{R}^{N\times D\times 1\times 1}$ is generated to indicate whether each channel is dropped or not, as shown in formula 1.
\begin{equation}m^i_{cd}=\left\{\begin{array}{ll}
0, & \text { if } f_i >\left(f_{\max } \cdot \lambda_{1}\right) \\
1 & \text { otherwise }
\end{array}\right.\end{equation}
\noindent where $m^i_{cd}$ equal to 0 means the $i$-th channel is dropped. The binary drop mask is then multiplied to input map $\boldsymbol{X}_3$ to get the channel-dropped feature map $\boldsymbol{X}_{cd}\in \boldsymbol{R}^{N\times D\times H\times W}$:

\begin{equation}
\boldsymbol{X}_{cd} = \boldsymbol{X}_3 \odot \boldsymbol{m}_{cd}
\end{equation}
\noindent where $\odot$ denotes broadcast element-wise multiplication.

\textbf{Spatial-Dropout}. A complementary symmetric structure is constructed for spatial dropout to induce the network to learn more comprehensive features. We first get the self-attention map $\boldsymbol{X}_a \in R^{N\times 1\times H\times W}$ via a channel-wise average pooling layer. Since the activation value of more discriminative areas in the attention map is higher, we set a threshold $\lambda_2$ to erase these areas to force the network to learn less discriminative features for classification but meaningful features for object localization, thereby avoiding location part domination. For the self-attention map $\boldsymbol{X}_a$, the maximum value of $i^{th}$ row is recorded as $g^{i}_{max}$. When the element $g^{ij}$ in row $i$ and column $j$ of the attention map is greater than the corresponding drop threshold $g^i_{max} \cdot \lambda_2$, the element is dropped; otherwise, the element is retained. As a result, we can obtain a binary spatial-dropout mask $\boldsymbol{m}_{sd} \in \boldsymbol{R}^{N\times 1\times H\times W}$:
\begin{equation}m^{ij}_{sd}=\left\{\begin{array}{ll}
0, & \text { if } g^{ij}>\left(g^i_{max} \cdot \lambda_2\right) \\
1 & \text { otherwise }
\end{array}\right.\end{equation}
\noindent where $m^{ij}_{sd}$ equal to 0 means the element in row $i$ and column $j$ of $\boldsymbol{X}_a$  should be discarded. When $\lambda_2$ decreases, more element values will be discarded.

However, when applying spatial-dropout throughout the training peroid, the most discriminative elements will always be ignored. As a result, the classification performance of the network will also be significantly reduced, which will also harm localization performance. In order to make up for the reduction of classification ability, we set up a reward branch to further enhance the discriminative elements. Specifically, the proposed network activates the self-attention map $\boldsymbol{X}_a$ through a sigmoid function to obtain an importance map $\boldsymbol{m}_{imp}$, where the intensity of each pixel in the importance map is close to 1 for the most discriminative features and close to 0 for the least discriminative features. During the training process, the network stochastically chooses either of the drop mask or importance map according to $drop\_rate$, and the selected one is merged into the input feature map $\boldsymbol{X}_{cd}$ to gain the spatial-dropped feature $\boldsymbol{X}_3^{*}\in R^{N\times D \times H \times W}$ by element-wise multiplication:

\begin{equation}\boldsymbol{X}_3^{*}=\left\{\begin{array}{ll}
\boldsymbol{X}_{cd} \odot \boldsymbol{m}_{sd}, & \text { if } (\alpha + drop\_rate) \textgreater 1 \\
\boldsymbol{X}_{cd} \odot \boldsymbol{m}_{imp} & \text { otherwise }
\end{array}\right.\end{equation}
\noindent where $\alpha$ is a random number from a uniform distribution on the interval [0, 1).

\begin{figure*}[!t]
	\centering{\includegraphics[scale=0.48]{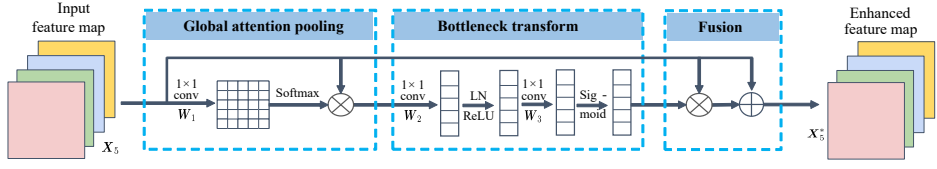}}
	\caption{Global context module. (1) Generate global context features. (2) Capture channel-wise dependencies. (3) Perform feature fusion to obtain enhanced feature maps.\label{fig:gcm}}
\end{figure*}

\subsection{Global Context Module}
The non-local strategy\cite{wang2018non} has been proved its efficiency on modeling long-distance dependencies of pixel pairs. NLNet \cite{wang2018non} learns a position-independent attention map for each position, which is time-consuming. SENet\cite{hu2018squeeze} uses the global context to calibrate the weights of different channels to adjust the channel dependence, in which the weight calibration inhabits its use of global context. In the paper, a new global context module like GCNet\cite{cao2019gcnet} is introduced to enhance the understanding of the global context in a more efficient way, as shown in Figure \ref{fig:gcm}.

The module can be divided into three stages: global attention pooling, bottleneck transform, and feature fusion. At the first stage, we obtain the attention weights via $1 \times 1$ convolutional layer  $\boldsymbol{W}_1$ and a softmax layer, and multiple it into the input to get the global context features $\boldsymbol{\beta}$, which is expected to help the model better understand the visual scene globally. After that, the model capture channel-wise dependencies through two $1 \times 1$ convolutional layers $\boldsymbol{W}_{2},\boldsymbol{W}_{3}$. In order to reduce the difficulty of model optimization, a layer normalization (LN) is inserted into the module (before ReLU). Sigmoid activation is also employed to learn a non-matually-exclusive relationship to improve original bottleneck transform of GCNet. Finally, unlike GCNet, which uses element-wise addition to fuse features, our enhanced features are fused into the original input by element-wise multiplication and then addition. 

Denoting $\boldsymbol{X}_5$ and $\boldsymbol{X}_5^*$ as the input and output feature map of the global context module,so this module can be formulated as :
\begin{equation}
\boldsymbol{X}_5^{*ij}=\boldsymbol{X}_5^{ij}+\boldsymbol{X}_5^{ij} \cdot \operatorname{Sig}\left(\boldsymbol{W}_{3} \operatorname{ReLU}\left(\operatorname{LN} \left(\boldsymbol{W}_{2}\boldsymbol{\beta}\right)\right)\right)
\end{equation}
%\begin{equation}
%\boldsymbol{\beta}=\sum_{n=1}^{N_{p}} \frac{e^{\boldsymbol{W}_{1} \boldsymbol{x}_{n}}}{\sum_{s=1}^{N_{p}} e^{\boldsymbol{W}_{1} \boldsymbol{x}_{s}}} \cdot \boldsymbol{x}_{n}
%\end{equation}

\begin{equation}
\boldsymbol{\beta}=\sum_{i=1}^{H} \sum_{j=1}^{W} \frac{e^{\boldsymbol{W}_{1} \boldsymbol{X}_{5}^{ij}}}{\sum_{s=1}^{H} \sum_{t=1}^{W} e^{\boldsymbol{W}_{1} \boldsymbol{X}_{5}^{st}}} \cdot \boldsymbol{X}_5^{ij}
\end{equation}

\noindent where $\operatorname{Sig}$ is the nonlinear activation function Sigmoid, $i$ and $j$ is the index of query positions, and $H$ and $W$ is the height and weight of the feature map. $\delta(\cdot)=\operatorname{Sig}\left(\boldsymbol{W}_{3} \operatorname{ReLU}\left(\mathrm{LN}\left(\boldsymbol{W}_{2}\boldsymbol{\beta}\right)\right)\right)$ indicates the bottleneck transform to capture channel-wise dependencies.

\begin{figure*}[!t]
	\centering{\includegraphics[scale=0.6]{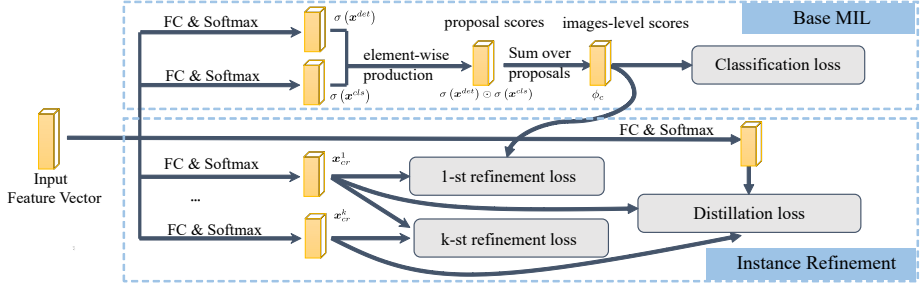}}
	\caption{ MIL and refinement module. (1) Train a basic mil classifier to roughly classify instances. (2) Build K instance classifiers and a distillation branch to optimize the output of the basic instance classifier.\label{fig:mil}}
\end{figure*}

\subsection{MIL and Refinement Module}
Following \cite{felipe2020distilling}, we build three submodules to classify and optimize instances, as shown in Figure \ref{fig:mil}. The first module trains a basic multiple instance learning network, which branches the proposal feature vectors into two streams to get $\boldmath{x}^{det}=\{\boldsymbol{x}^{det}_1,\boldsymbol{x}^{det}_2,\cdots,\boldsymbol{x}^{det}_{|R|}\},\boldsymbol{x}^{cls}=\{\boldsymbol{x}^{cls}_1,\boldsymbol{x}^{cls}_2,\cdots,\boldsymbol{x}^{cls}_{|R|}\}$ by two FC layers, where $|R|$ is the number of proposals. These two matrices are passed through a softmax layer on the classes and proposals dimensions respectively to get the activated prediction result $\sigma\left(\boldsymbol{x}^{det}\right), \sigma\left(\boldsymbol{x}^{cls}\right)$ . The predictions of the two branches are fused by element-wise multiplication to obtain the category prediction score $\boldsymbol{x}_{r}=\sigma\left(\boldsymbol{x}_r^{det}\right) \odot \sigma\left(\boldsymbol{x}_r^{cls}\right)$ of the $r^{th}$ region proposal. Finally, the prediction score $\phi_{c}=\sum_{r=1}^{|R|} \boldsymbol{x}_{cr}$ of this image for class $c$ can be obtained by summing up the scores over proposal dimensions. We use multi-classes cross-entropy loss to guide the training of this submodule:
\begin{equation}
\mathcal{L}_{cls}=-\sum_{c=1}^{C}\left\{y_{c} \log \phi_{c}+\left(1-y_{c}\right) \log \left(1-\phi_{c}\right)\right\}
\end{equation}
\noindent where $y_{c}=1$ indicates that the input image contain $c^{th}$ class object and $y_{c}=0$ otherwise.

The second submodule builds K instance classifiers to optimize the output of the basic instance classifier. Each classifier is implemented by a FC layer and a softmax layer along $C+1$ categories (background is considered as $0^{th}$class). The output of the $k^{th}$ classifier is considered as the  supervision information of the $(k+1)^{th}$ classifier. so we can train the $k^{th}$ refined instance classifier based on the loss function $\mathcal{L}_{ref}^k$ in formula 8.
\begin{equation}
\mathcal{L}_{ref}^{k}=-\frac{1}{|R|} \sum_{r=1}^{|R|} \sum_{c=1}^{C+1} w_{r}^{k} y_{c r}^{k} \log x_{c r}^{k}
\end{equation}

\noindent where $w_{r}^{k}$ is the loss weight term of $r^{th}$ region proposal to reduce the noise of supervision and the same as \cite{tang2017multiple}. $y_{c r}^{k}$ is the pseudo grouth truth information for class $c$ from the $(k-1)^{th}$ classifier, and $x_{c r}^{R k}$ indicates the prediction score of $r^{th}$ region proposal for class $c$ in the $k^{th}$ refinement branch.

Finally, we average the outputs of the K refinements agents outputs as the supervision to guide the distillation agent, which consist of a single FC layer and a softmax layer over class dimension. The distillation loss $\mathcal{L}_{dis}$ is the same as the refinement loss $\mathcal{L}_{ref}$. As a result, The final loss function of the entire network is as follows:
\begin{equation}
\mathcal{L} = \mathcal{L}_{cls} + \mathcal{L}_{dis} + \sum^K_{k=1} \mathcal{L}^k_{ref}.
\end{equation}

%-------------------
\section{Experiments}\label{sec4}
In this section, we will first introduce the dataset we used and the implementation details of our proposed approach. Then we will explore the contributions of each proposed module by the ablation experiments. Finally, we will compare the performance of our proposed network with the-state-of-art methods.

\subsection{Datasets and Evaluation Metrics}
we evaluate our method on the challenging PASCAL VOC2007 datasets\cite{everingham2010pascal} which have 9963 images for 20 object classes and are divided into three subsets: training, verification, and testing sets. The trainval set (5011 images) is chosen to train our proposed network. As we focus on weakly-supervised detection, only image-level labels (presence or absence of a class in the chosen image)are utilized during training. For testing, we evaluating our network using mean average precision (mAP)\cite{everingham2010pascal} and correct localization (CorLoc) metrics\cite{deselaers2012weakly}. All these metrics are based on the PASCAL criteria of IoU \textgreater \,0.5 between ground truths and predicted boxes.

\subsection{Implementation Details}
We use Selective search\cite{uijlings2013selective} to generate original region proposal and build our proposed network on VGG16\cite{simonyan2014very} pre-trained on  ImageNet\cite{deng2009imagenet}. We add the CADM module to the back layer of pooling3. The last max-pooling layer is replaced by ROI pooling and the last FC layer and softmax loss layer are replaced by the layer described in Section 3.3. We insert the global context module in front of the ROI layer.

The original input images are resized into five different scales \{480,576,688,864,1200\} concerning the smallest image dimension. The resized scale of a training image is randomly selected and the image is randomly horizontal flipped. In this way, each image is augmented into a total of ten images as many other WSOD methods do\cite{tang2017multiple,tang2018pcl,wan2018min,wan2019c}. During network learning, we employ the SGD algorithm with momentum 0.9, weight decay $5 \times 10^{-4}$ and batch size 4. The model iterates 50K iterations, where the learning rate is set to $5 \times 10^{-4}$ for the first 20K iterations and then decreases to  $5 \times 10^{-5}$ in the following 30K iterations. During testing, all ten augmented images are passed into the network, and take the averages as the final outputs. Non-maximum suppression is applied to all prediction with the IoU threshlod set to 0.3. Our experiments are implemented based on PyTorch deep learning framework and a NVIDIA GTX 1080Ti GPU.

\begin{table}[!t]
	\caption{Ablation study : AP performance(\%) of each category on the PASCAL VOC 2007 test set}
	\label{tab_aba_map}
	\centering
	\begin{tabular}{lcccccccccc}
		\hline\noalign{\smallskip}
		Methods                      & aero & bike & bird & boat & bottle & bus  & car  & cat  & chair & cow   \\
		\noalign{\smallskip}\hline\noalign{\smallskip}
		Baseline\cite{felipe2020distilling}     		& 63.1 & \textbf{66.4} & 46.1 & 25.4 & 16.9   & 70.8 & 68.8 & 53.2 & 14.9  & 56.8  \\
		 + CADM       & 64.6 & 64.7 & 53.1  & \textbf{33.1} & \textbf{23.2} & 70.4 & \textbf{70.6} & 22.2 & \textbf{22.1} & 64.1	  \\
		 + GCM       	&66.0 & 66.0 & \textbf{56.8}  & 20.3 & 19.1 & \textbf{72.5} & \textbf{70.6} & 59.5 & 19.6 & 64.3   \\
		 + Both   & \textbf{66.5} & 65.6 & 56.5 & 26.8 & 19.7 & 69.9 & 69.0 & \textbf{61.3} & 21.5 & \textbf{66.9}   \\

		\hline\noalign{\smallskip}
%		\hline\noalign{\smallskip}
		Methods								   & table & dog  & horse & mbike & person & plant & sheep & sofa & train & tv  \\
		\noalign{\smallskip}\hline\noalign{\smallskip}
		Baseline\cite{felipe2020distilling}       	   & 41.5  & \textbf{53.7} & 42.7  & 70.0  & \textbf{2.9}   & \textbf{20.6}  & 42.8  & 44.8 & 50.8  & 68.3   \\
		 + CADM            & \textbf{46.2} & 27.1 & \textbf{49.3} & \textbf{70.8} & 2.6 & 19.5 & \textbf{57.1} & \textbf{55.3} & 64.6 & 69.9  \\
		 + GCM             & 42.8 & 47.8 & 42.8 & 68.5 & 2.5 & 20.0 & 47.6 & 48.7 & 62.9 & 64.2  \\
		 + Both   & 43.1 & 50.4 & 49.0 & 70.1 & 2.3 & 20.1 & 53.9 & 47.4 & \textbf{65.7} & \textbf{70.7}  \\

		\noalign{\smallskip}\hline
		
	\end{tabular}
\end{table}

\begin{table}[!t]
	\caption{Ablation study : CorLoc performance(\%) of each category on the PASCAL VOC 2007 trainval set}
	\label{tab_aba_corloc}
	\centering
	%\scalebox{0.9}{
	\begin{tabular}{lcccccccccc}
		\hline\noalign{\smallskip}
		Methods     						                & aero & bike & bird & boat & bottle & bus  & car  & cat  & chair & cow   \\
		\noalign{\smallskip}\hline\noalign{\smallskip}
		Baseline\cite{felipe2020distilling}     & \textbf{84.6} & 78.4 & 59.2 & 49.5 & 44.7   & 77.7 & 85.0 & 61.0 & 34.8  & 75.3  \\
		 + CADM    & \textbf{84.6} & 77.6 & 70.6 & 56.4 & \textbf{50.4}  & 78.7 & 84.4 & 39.0 & \textbf{47.9} & 82.9	  \\
		 + GCM     & 84.2 & \textbf{82.4} & 70.3 & \textbf{64.4} & 44.7 & \textbf{79.7} & \textbf{86.6} & 47.1 & 46.9 & \textbf{84.2}   \\
		 + Both    & 84.2 & 75.7 & \textbf{75.1} & 50.0 & 40.8 & 77.7 & 82.3 & \textbf{68.6} & 46.3 & 82.2   \\
		
		\hline\noalign{\smallskip}
		%		\hline\noalign{\smallskip}
		Methods								   & table & dog  & horse & mbike & person & plant & sheep & sofa & train & tv  \\
		\noalign{\smallskip}\hline\noalign{\smallskip}
		Baseline\cite{felipe2020distilling}    & 44.1  &\textbf{ 70.5}   & 65.0  & 88.8  & \textbf{11.4}   & \textbf{57.1}  & 73.2 & 51.9 & 66.2  & \textbf{82.4}   \\
		 + CADM       & 52.1  & 38.4 & 70.4  & \textbf{90.4}  & 10.8  & 55.7 & 81.4 & \textbf{69.4} & \textbf{78.7}  & 82.1  \\
 		 + GCM        & 48.7 & 40.0 & \textbf{73.5} & 88.8 & 11.0 & 55.7 & \textbf{82.5} & 66.4 & 70.3 & \textbf{82.4}  \\
		 + Both       & \textbf{52.9} & 63.5 & 73.1 & 89.6 & 10.6 & 52.7 & 79.4 & 55.4 & \textbf{78.7} & 82.1  \\
		
		\noalign{\smallskip}\hline
		
	\end{tabular}
%}
\end{table}

\begin{table}[t]
	\caption{Ablation study : average detection and localization performance(\%) on PASCAL VOC 2007}
	\label{tab_aba_map+corloc}       % Give a unique label
	\centering
	\begin{tabular}{lcc}
		\hline\noalign{\smallskip}
		Methods  & mAP & CorLoc   \\
		\noalign{\smallskip}\hline\noalign{\smallskip}
		Baseline\cite{felipe2020distilling} & 46.0  & 63.0  \\
		 + CADM  & 47.5 & 65.1    \\
		 + GCM   & 48.1 & 65.5      \\
		 + CADM + GCM  & \textbf{49.8} & \textbf{66.0}    \\
		\noalign{\smallskip}\hline
	\end{tabular}
\end{table}

\begin{table}[t]
	\caption{Ablation study : different way of feature fusion in GCM on PASCAL VOC 2007.  $\checkmark$ indicates that the component is used.}
	\label{tbl:gcb_compoment}       % Give a unique label
	\centering
	\begin{tabular}{llll}
		\hline\noalign{\smallskip}
		multiplication & addition & multiplication + addition &  mAP   \\
		\noalign{\smallskip}\hline\noalign{\smallskip}
		&   &  & 46 \\
		\multicolumn{1}{c}{$\checkmark$} &  &  & 47.8  \\
		& \multicolumn{1}{c}{$\checkmark$} &   & 47.0 \\
		&  & \multicolumn{1}{c}{$\checkmark$} & \textbf{48.1}  \\
		\noalign{\smallskip}\hline
	\end{tabular}
\end{table}

\begin{table}[!t]
	\caption{Ablation study : different dimension of dropout in CADM on PASCAL VOC 2007.  $\checkmark$ indicates that the component is used.}
	\label{tbl:cadb_compoment}       % Give a unique label
	\centering
	\begin{tabular}{llll}
		\hline\noalign{\smallskip}
		channel-dropout & spatial-dropout & channel + spatial dropout &  mAP   \\
		\noalign{\smallskip}\hline\noalign{\smallskip}
		&   &  & 48.1 \\
		\multicolumn{1}{c}{$\checkmark$}   &  &  & 47.6  \\
		& \multicolumn{1}{c}{$\checkmark$} &  & 48.8 \\
		&  & \multicolumn{1}{c}{$\checkmark$} & \textbf{49.8}  \\
		\noalign{\smallskip}\hline
	\end{tabular}
\end{table}

\subsection{Ablation experiments}
we conduct extensive ablation experiments on PASCAL VOC2007 to prove the effectiveness of our proposed network and respectively validate the contribution of each component including CADM and GCM.

\subsubsection{Baseline}\label{subsec4.3.1}
We use Boosted-OICR\cite{felipe2020distilling} as our baseline. We re-run the author's code multiple times with the same configuration, but the highest mAP we can achieve is only 46.0\% due to different cuda versions and GPUs. \footnote{More details about the issue can refer to {https://github.com/ppengtang/pcl.pytorch/issues/9}}

\subsubsection{GCM}
We conduct experiments with and without GCM to illustrate the effectiveness of GCM and denote the network with GCM as \textbf{+GCM}, which does not include CADM and report the results in Table \ref{tab_aba_map} , Table \ref{tab_aba_corloc} and Table \ref{tab_aba_map+corloc}. The detection performance of 13 classes and the localization performance of 11 classes have been improved. Figure \ref{tab_aba_map+corloc} shows that applying global context module to our proposed network could improve the performance of the model by at least 2.1\% mAP and 2.5\% CorLoc.

We also explore the effect of different way of feature fusion on model performance on PASCAL VOC 2007 and the results are shown in Table \ref{tbl:gcb_compoment}. It shows that multiplication and then addition is more effective than simple multiplication or addition in the fusion stage.

\begin{figure*}[t]
	\centering{\includegraphics[scale=0.4]{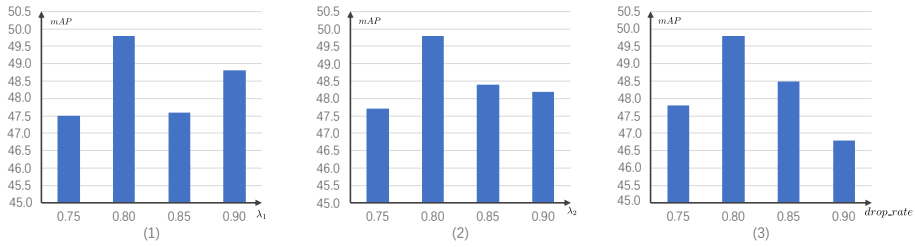}}
	\caption{Ablation study : influence of different values of $\lambda_1$ (left), $\lambda_2$ (middle), $drop\_rate$ (right) in CADB on model performance.}
	\label{fig:aba_cadb}
\end{figure*}

\begin{table}[t]
	\caption{Detection performance (\%) of each category on the VOC 2007 test set. Comparison to the state-of-the-arts.}
	\label{tab_voc_2007_detection}
	\centering
	\begin{tabular}{llccccccccc}
		\hline\noalign{\smallskip}
		Methods                      & aero & bike & bird & boat & bottle & bus  & car  & cat  & chair & cow   \\
		\noalign{\smallskip}\hline\noalign{\smallskip}
		WSDDN\cite{bilen2016weakly}      & 46.4 & 58.3 & 35.5 & 25.9 & 14.0   & 66.7 & 53.0 & 39.2 & 8.9   & 41.8  \\
		DSTL\cite{jie2017deep}           & 52.2 & 47.1 & 35	& 26.7 &15.4    &61.3  & 66	  & 54.3 & 3	 & 53.6	  \\
		OICR\cite{tang2017multiple}      & 58.0 & 62.4 & 31.1 & 19.4 & 13.0   & 65.1 & 62.2 & 28.4 & 24.8  & 44.7     \\
		WCCN\cite{diba2017weakly}     & 49.5 & 60.6 & 38.6 & 29.2 & 16.2   & \textbf{70.8} & 56.9 & 42.5 & 10.9  & 44.1   \\
		PCL\cite{tang2018pcl}     & 54.4 & 69.0 & 39.3 & 19.2 & 15.7   & 62.9 & 64.4 & 30.0   & \textbf{25.1}  & 52.5   \\
		TS2C\cite{wei2018ts2c}    & 59.3 & 57.5 & 43.7 & 27.3 & 13.5   & 63.9 & 61.7 & 59.9 & 24.1  & 46.9   \\
		C-WSL\cite{gao2018c}              & 62.7 & 63.7 & 40.0 & 25.5 & 17.7   & 70.1 &68.3  & 38.9 & 25.4 & 54.5   \\
		W2F\cite{zhang2018w2f}       & 60.9 & 68.7 & 47.1 & 31.7 & 14.2   & 71.2 & 68.9 &24.5 &23.5 &57.6   \\
		WeakRPN\cite{tang2018weakly}  & 57.9 & \textbf{70.5} & 37.8 & 5.7 & \textbf{21.0} & 66.1 & \textbf{69.2} & 59.4 & 3.4   & 57.1   \\
		BOICR\cite{felipe2020distilling} & 63.1 & 66.4 & 46.1 & 25.4 & 16.9   & \textbf{70.8} & 68.8 & 53.2   & 14.9  & 56.8    \\
		MELM\cite{wan2018min}             & 55.6 & 66.9 & 34.2 & 29.1 & 16.4   & 68.8 & 68.1 & 43.0   & 25.0    & 65.6   \\
		ZLDN\cite{zhang2018zigzag}        & 55.4 & 68.5 & 50.1 & 16.8 & 20.8 & 62.7 & 66.8 & 56.5 & 2.1 & 57.8   \\
		WSCDN\cite{wang2018collaborative} & 61.2 & 66.6 & 48.3 & 26.0 & 15.8 & 66.5 & 65.4 & 53.9 & 24.7 & 61.2   \\
		C-MIL\cite{wan2019c}              & 62.5 & 58.4 & 49.5 & \textbf{32.1} & 19.8   & 70.5 & 66.1 & \textbf{63.4} & 20.0  & 60.5  \\	
		
		Ours & \textbf{66.5} & 65.6 & \textbf{56.45} & 26.8 & 19.7 & 69.9 & 69.0 & 61.3 & 21.5 & \textbf{66.9}  \\
		
		\hline\noalign{\smallskip}
		%		\hline\noalign{\smallskip}
		Methods	& table & dog  & horse & mbike & person & plant & sheep & sofa & train & tv  \\
		\noalign{\smallskip}\hline\noalign{\smallskip}
		WSDDN\cite{bilen2016weakly}       & 26.6  & 38.6 & 44.7  & 59.0  & 10.8  & 17.3  & 40.7  & 49.6 & 56.9  & 50.8  \\
		DSTL\cite{jie2017deep}           & 24.7	 & 43.6 & 48.4  & 65.8	& 6.6	& 18.8	& 51.9	& 43.6 & 53.6  & 62.4  \\
		OICR\cite{tang2017multiple}      & 30.6  & 25.3 & 37.8  & 65.5  & 15.7  & 24.1  & 41.7  & 46.9 & 64.3  & 62.6  \\
		WCCN\cite{diba2017weakly}        & 29.9  & 42.2 & 47.9  & 64.1  & 13.8  & 23.5  & 45.9  & 54.1 & 60.8  & 54.5  \\
		PCL\cite{tang2018pcl}           & 44.4  & 19.6 & 39.3  & 67.7  & 17.8  & 22.9  & 46.6  & 57.5 & 58.6  & 63.0  \\
		TS2C\cite{wei2018ts2c}           & 36.7  & 45.6 & 39.9  & 62.6  & 10.3  & 23.6  & 41.7  & 52.4 & 58.7  & 56.6  \\
		C-WSL\cite{gao2018c}              & 41.6  & 29.9 & 37.9  & 64.2  & 11.3  & 27.4  & 49.3  & 54.7 & 61.4  & 67.4  \\
		W2F\cite{zhang2018w2f}          & 43.6  & 20.9 & 47.9  & 66.0  & 11.3  & 22.3  & 56.4  & 57.7 & 61.1  & 60.1  \\
		WeakRPN\cite{tang2018weakly}        & \textbf{57.3} & 35.2 & \textbf{64.2}  & 68.6 & \textbf{32.8} & \textbf{28.6} & 50.8 & 49.5 & 41.1  & 30.0    \\
		BOICR\cite{felipe2020distilling}  & 41.5  & \textbf{53.7} & 42.7  & 70  & 2.9   & 20.6  & 42.8  & 44.8 & 50.8  & 68.3    \\
		MELM\cite{wan2018min}            & 45.3  & 53.2 & 49.6  & 68.6  & 2.0   & 25.4  & 52.5  & 56.8 & 62.1  & 57.1  \\
		ZLDN\cite{zhang2018zigzag}       & 47.5 & 40.1 & 69.7 & 68.2 & 21.6 & 27.2 & 53.4 & 56.1 & 52.5 & 58.2  \\
		WSCDN\cite{wang2018collaborative} & 46.2 & 53.5 & 48.5 & 66.1 & 12.1 & 22.0 & 49.2 & 53.2 & \textbf{66.2} & 59.4  \\
		C-MIL\cite{wan2019c}              & 52.9  & 53.5 & 57.4  & 68.9  & 8.4 & 24.6  & 51.8 & \textbf{58.7} & \textbf{66.7}  & 63.5  \\	
		
		Ours & 43.1 & 50.4 & 49.0 & \textbf{70.1} & 2.3 & 20.1 & \textbf{53.9} & 47.4 & 65.7 & \textbf{70.7} \\
		\noalign{\smallskip}\hline
		
	\end{tabular}
\end{table}

\begin{table}[t]
	\caption{Localization performance (\%) of each category on the VOC 2007 trainval set. Comparison to the state-of-the-arts.}
	\label{tab_voc_2007_corloc}
	\centering
	\begin{tabular}{llccccccccc}
		\hline\noalign{\smallskip}
		Methods                          & aero & bike & bird & boat & bottle & bus  & car  & cat  & chair & cow  \\
		\noalign{\smallskip}\hline\noalign{\smallskip}
		WSDDN\cite{bilen2016weakly}       & 65.1 & 58.8 & 58.5 & 33.1 & 39.8   & 68.3 & 60.2 & 59.6 & 34.8  & 64.5   \\
		DSTL\cite{jie2017deep}           & 72.7 & 55.3 & 53.0 & 27.8 & 35.2   & 68.6 & 81.9 & 60.7 & 11.6  & 71.6   \\
		OICR\cite{tang2017multiple}      & 81.7 & 80.4 & 48.7 & 49.5 & 32.8   & 81.7 & 85.4 & 40.1 & 40.6  & 79.5   \\
		WCCN\cite{diba2017weakly}        & 83.9 & 72.8 & 64.5 & 44.1 & 40.1   & 65.7 & 82.5 & 58.9 & 33.7  & 72.5   \\
		PCL\cite{tang2018pcl}           & 79.6 & \textbf{85.5} & 62.2 & 47.9 & 37.0 & \textbf{83.8} & 83.4 & 43.0   & 38.3  & 80.1 \\
		TS2C\cite{wei2018ts2c}           & 84.2 & 74.1 & 61.3 & \textbf{52.1} & 32.1   & 76.7 & 82.9 & 66.6 & 42.3  & 70.6 \\
		C-WSL\cite{gao2018c}              & 86.3 & 80.4 & 58.3 & 50.0 & 36.6 &85.8 &86.2 &47.1 &42.7 &81.5 \\
		WeakRPN\cite{tang2018weakly}        & 77.5 & 81.2 & 55.3 & 19.7 & 44.3   & 80.2 & \textbf{86.6} & \textbf{69.5} & 10.1  & \textbf{87.7}   \\
		BOICR\cite{felipe2020distilling}  & 84.6 & 78.4 & 59.2 & 49.5 & 44.7 & 77.7 & 85.0 & 61.0 & 34.8  & 75.3     \\
		ZLDN\cite{zhang2018zigzag}       & 74.0 & 77.8 & 65.2 & 37.0 & \textbf{46.7} & 75.8 &83.7 &58.8 &17.5 &73.1 \\
		WSCDN\cite{wang2018collaborative} & 85.8 & 80.4 & 73.0 & 42.6 & 36.6 & 79.7 & 82.8 & 66.0 & 34.1 & 78.1   \\
		ours &\textbf{84.2} & 75.7 & \textbf{75.1} &50.0 & 40.8 & 77.7 & 82.3 & 68.6 & \textbf{46.3} & 82.2  \\
		
		\hline\noalign{\smallskip}
		
		Methods & table & dog  & horse & mbike & person & plant & sheep & sofa & train & tv \\
		\noalign{\smallskip}\hline\noalign{\smallskip}
		WSDDN\cite{bilen2016weakly}       & 30.5  & 43.0 & 56.8  & 82.4  & 25.5   & 41.6  & 61.5  & 55.9 & 65.9  & 63.7  \\
		DSTL\cite{jie2017deep}           & 29.7  & 54.3 & 64.3  & 88.2  & 22.2   & 53.7  & 72.2  & 52.6 & 68.9  & 75.5   \\
		OICR\cite{tang2017multiple}      & 35.7  & 33.7 & 60.5  & 88.8  & 21.8   & 57.9  & 76.3  & 59.9 & 75.3  & 81.4     \\
		WCCN\cite{diba2017weakly}        & 25.6  & 53.7 & 67.4  & 77.4  & 26.8   & 49.1  & 68.1  & 27.9 & 64.5  & 55.7   \\
		PCL\cite{tang2018pcl}           & 50.6  & 30.9 & 57.8  & 90.8 & 27.0 & 58.2  & 75.3  & \textbf{68.5} & 75.7  & 78.9  \\
		TS2C\cite{wei2018ts2c}           & 39.5  & 57.0 & 61.2  & 88.4  & 9.3    & 54.6  & 72.2  & 60.0   & 65.0    & 70.3   \\
		C-WSL\cite{gao2018c}              &42.2 &42.6 &50.7 &90.0 &14.3 &61.9 &85.6 &64.2 &77.2 &\textbf{82.4}   \\
		WeakRPN\cite{tang2018weakly}        & \textbf{68.4}  & 52.1 & \textbf{84.4}  & \textbf{91.6}  & \textbf{57.4}   & \textbf{63.4}  & \textbf{77.3}  & 58.1 & 57.0 & 53.8  \\
		BOICR\cite{felipe2020distilling}  & 44.1  & \textbf{70.5}   & 65.0  & 88.8  & 11.4   & 57.1  & 73.2 & 51.9 & 66.2  & \textbf{82.4}     \\
		ZLDN\cite{zhang2018zigzag}       &49.0 &51.3 &76.7 &87.4 & 30.6 &47.8 &75.0 &62.5 &64.8 &68.8   \\
		WSCDN\cite{wang2018collaborative} & 36.9 & 68.6 & 72.4 & \textbf{91.6} & 22.2 & 51.3 & 79.4 & 63.7 & 74.5 & 74.6   \\
		
		Ours & 52.9 & 63.5 & 73.1 & 89.6 & 10.6 & 52.7 & 79.4 & 55.4 & \textbf{78.7} & 82.1 \\
		\noalign{\smallskip}\hline
		
	\end{tabular}
\end{table}

\subsubsection{CADM}
To verify the effectiveness of CADM, we also conduct experiments with and without CADM. We empirically set original threshold $\lambda_1,\lambda_2$ to 0.8, $drop\_rate$ to 0.8 and denote the network with GAM as \textbf{+CADM}, which does not include a global context module. From Table \ref{tab_aba_map} , Table \ref{tab_aba_corloc} and Table \ref{tab_aba_map+corloc}, we can conclude that CADM does help the detector learn comprehensive features rather than only discriminative features and brings at least 1.5\% mAP and 2.1\% CorLoc improvement. 

We also further investigate the effect of different $\lambda_1,\lambda_2,drop\_rate$ on detection performance, as shown in Figure \ref{fig:aba_cadb}. When exploring the role of a certain value, the other two values are fixed at 0.8. From the three subgraphs, we can find that different values have a greater impact on the performance of the model, sometimes it will greatly improve the performance of the model (49.8\% vs 48.1\%), but sometimes it will hurt the performance of the model (46.8\% vs 48.1\%). As a result, an appropriate threshold selection is very important.

From Table \ref{tbl:cadb_compoment}, we can find that using the spatial-dropout submodule can slightly improve the performance of the model, but simply using the channel-dropout submodule will damage the performance of the model to a certain extent. Interestingly, when both submodules are inserted into the network in a cascaded way, the performance of the model is greatly improved. We believe that this is because a simple channel-dropout submodule may throw away too many elements, and the increase in localization performance cannot make up for the loss of classification performance, thereby compromising the final accuracy. But when the two are combined, the more discriminative features are moderately discarded to induce the network to learn those less discriminative features, thereby greatly improving the performance of the model.

\begin{table}[t]
	\caption{Average detection and localization performance on PASCAL VOC 2007.}
	\label{tab_voc_2007_detection+loc}
	\centering
	\begin{tabular}{lcc}
		\hline\noalign{\smallskip}
		Methods & mAP & CorLoc\\
		\noalign{\smallskip}\hline\noalign{\smallskip}
		
		WSDDN \cite{bilen2016weakly} & 39.2 & 53.5 \\
		DSTL \cite{jie2017deep} & 41.7 & 56.1 \\
		OICR \cite{tang2017multiple} &42.0 & 60.6   \\
		WCCN \cite{diba2017weakly}  & 42.8  & 56.7  \\
		PCL \cite{tang2018pcl} & 43.5  & 62.7 \\
		TS2C \cite{wei2018ts2c} & 44.3  & 61.0 \\
		C-WSL \cite{gao2018c} & 45.6  & 63.3 \\
		WSD+FSD1 \cite{zhang2018w2f} & 45.8  & 65.0 \\
		WeakRPN \cite{tang2018weakly} & 45.3  & 63.8 \\
		BOICR \cite{felipe2020distilling} & 46.0  & 63.0 \\
		MELM \cite{wan2018min} &  47.3 & 61.4 \\
		ZLDN \cite{zhang2018zigzag} & 47.3  & 61.2 \\
		WSCDN \cite{wang2018collaborative} & 48.3  & 64.7 \\
		C-MIL \cite{wan2019c}  & \textbf{50.3}  & 65.0 \\
		Ours & 49.8 & \textbf{66.0} \\
		\noalign{\smallskip}\hline
	\end{tabular}
\end{table}

\begin{table}[t]
	\caption{Average detection and localization performance on PASCAL VOC 2012.}
	\label{tab_voc_2012_detection+loc}
	\centering
	\begin{tabular}{lcc}
		\hline\noalign{\smallskip}
		Methods & mAP & CorLoc\\
		\noalign{\smallskip}\hline\noalign{\smallskip}
		DSTL \cite{jie2017deep} & 38.3 & 58.8 \\
		OICR \cite{tang2017multiple} &37.9 & 62.1   \\
		WCCN \cite{diba2017weakly}  & 37.9  & -  \\
		PCL \cite{tang2018pcl} & 40.6  & 63.2 \\
		TS2C \cite{wei2018ts2c} & 40.0  & 64.4 \\
		C-WSL \cite{gao2018c} & 41.5  & 64.2 \\
		WSD+FSD1 \cite{zhang2018w2f} & 42.4  & 65.5 \\
		WeakRPN \cite{tang2018weakly} & 40.8  & 64.9 \\
		MELM \cite{wan2018min} &  42.4 & - \\
		ZLDN \cite{zhang2018zigzag} & 42.9  & 61.5 \\
		WSCDN \cite{wang2018collaborative} & 43.3  & 65.2 \\
		C-MIL \cite{wan2019c}  & 46.6  & 67.4 \\
		SDCN \cite{li2019weakly}  & 43.5  & 67.9 \\
		BOICR \cite{felipe2020distilling} & 46.7  & 66.3 \\
		MIL-OICR+GAM+REG \cite{yang2019towards} & 46.8  & \textbf{69.5} \\
		Ours & \textbf{46.9} & 66.5 \\
		\noalign{\smallskip}\hline
	\end{tabular}
\end{table}

\subsection{Comparison with state-of-the-art}
Table \ref{tab_voc_2007_detection} and Table \ref{tab_voc_2007_detection+loc} shows the detection performance of our proposed method and a comparison with other state-of-the-art methods on the PASCAL VOC 2007 test set. It can be seen that our proposed method improves the original Boosted-OICR\cite{felipe2020distilling} code in 3.8{\%} mAP, and outperformed WSDDN\cite{bilen2016weakly}, OICR\cite{tang2017multiple}, WSCDN\cite{wang2018collaborative}, MELM\cite{wan2018min} and other approaches by 1.5{\%} $\sim$ 10.6{\%}. Meanwhile, our method achieved the highest AP results in 6 classes(aeroplane, bird, motorbike, cow, sheep and tv).

We also evaluated object location performance of our network  and compared it with other SOTA on the PASCAL VOC 2007 trainval set in Table \ref{tab_voc_2007_corloc} and Table \ref{tab_voc_2007_detection+loc}. The proposed network reached the best localization performance in 4 of the 20 classes(aeroplane, bird, chair and train). our network respectively outperformed WSDDN\cite{bilen2016weakly}, OICR\cite{tang2017multiple}, WSCDN\cite{wang2018collaborative}, MELM\cite{wan2018min}, C-MIL\cite{wan2019c} and other approaches by 1.0{\%} $\sim$ 12.5{\%}.

We achieve a performance of 46.9\% mAP and 66.5\% CorLoc on Pascal VOC 2012, which is supervior to previous work in with gain of about 0.1{\%} $\sim$ 8.6{\%} mAP in Table \ref{tab_voc_2012_detection+loc}.

\begin{figure*} [!ht]
	\centering{\includegraphics[scale=0.75]{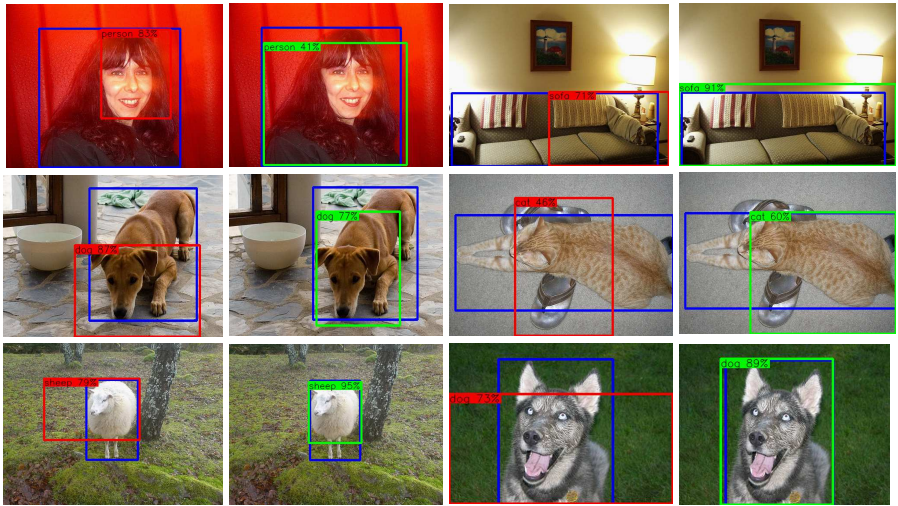}}
	\caption{Visualization of our detection results(2,4 columns) and baseline\cite{felipe2020distilling} detection results(1,3 columns). The bule, green and red boxes repectively indicies the ground-truth boxes, correct detctions(IoU \textgreater \,0.5 with grouth-truth boxes ) and wrong detections. The label in each box is the category prediction and its confidence score.}
	\label{fig:Visualization}
\end{figure*}

\subsection{Visualization}
The visualization of our network and baseline\cite{felipe2020distilling} detection results is shown in Figure \ref{fig:Visualization}. The first and third columns are the detection results of \cite{felipe2020distilling}, and the other two columns are our detection results. Our proposed methods have achieved better detection results than \cite{felipe2020distilling}, and the part-dominated problem has been better alleviated as shown in the first two rows. As can be seen from the third row of pictures, our method can also improve the problem of too large a prediction box.

%--------------------------
\section{Conclusion}
In this paper, we propose a simple but effective architecture for weakly supervised object detection. The network selectively discards the most discriminative features in the channel and spatial dimensions based on attention mechanism. To understand the image context information better, a global context module is also introduced into MIL. We have conducted extensive experiments and results show substantial and distinctive improvement of the our proposed method.

%--------------------------
\begin{acknowledgements}
	This work is supported by the National Natural Science Foundation of China(grant no. 61573168)
\end{acknowledgements}

\bibliographystyle{spmpsci}      % mathematics and physical sciences

\bibliography{cadm}

\end{document}